\documentclass[sigconf,nonacm]{acmart}

\usepackage{underscore}
\usepackage{microtype}
\usepackage{algorithm}
\usepackage{algpseudocode}
\usepackage{enumitem}
\usepackage{pifont}
\usepackage{subcaption}
\usepackage{multirow}
\usepackage{tikz}
\usetikzlibrary{arrows.meta, positioning, calc}
\definecolor{rpblue}{HTML}{4878A8}
\definecolor{rpred}{HTML}{9E4A4A} 
\graphicspath{{figures/}}

\AtBeginDocument{%
  }

\newcommand{\fullsupport}{\ensuremath{\bigcirc}}
\newcommand{\nosupport}{\ding{55}}

\setcopyright{none}
\renewcommand\footnotetextcopyrightpermission[1]{}

\AtBeginDocument{\setlength{\bibsep}{4pt plus 1pt minus 1pt}}

\begin{document}

\title{\mbox{ros2probe: Non-intrusive, Kernel-selective Observability} \mbox{for Robot Operating System~2 Middleware}}

\author{Jisang Yu}
\authornote{Both authors contributed equally to this research.}
\affiliation{%
  \institution{DGIST}
  \city{Daegu}
  \country{Republic of Korea}}
\email{julienyu@dgist.ac.kr}

\author{Sanghoon Lee}
\authornotemark[1]
\affiliation{%
  \institution{DGIST}
  \city{Daegu}
  \country{Republic of Korea}}
\email{leesh2913@dgist.ac.kr}

\author{Yeonwoo Choi}
\affiliation{%
  \institution{DGIST}
  \city{Daegu}
  \country{Republic of Korea}}
\email{cyw040306@dgist.ac.kr}

\author{Kyung-Joon Park}
\affiliation{%
  \institution{DGIST}
  \city{Daegu}
  \country{Republic of Korea}}
\email{kjp@dgist.ac.kr}

\renewcommand{\shortauthors}{Yu et al.}

\begin{abstract}
\looseness=-1 Robot Operating System 2~(ROS~2), the de facto standard middleware framework for robots, runs each robot as a graph of nodes communicating over the Data Distribution Service~(DDS), a publish/subscribe substrate.
Observing this inter-node communication in real time is essential to robot development, yet observing it has a price.
A tool can receive data only by joining the DDS domain as a subscriber that discovery has matched to the publisher, so the act of observing folds the tool into the system it measures and perturbs it.
We define this protocol-inherent perturbation as the observer's probe effect.
It inflates the discovery plane, adds deserialization cost on the observer, makes the loss it reports diverge from what the subscriber actually received, and near saturation displaces the subscriber's own messages.
The only existing escape, capturing all wire traffic passively, discards the ROS~2 message semantics and scales with total traffic, not with what is observed.
We present ros2probe, a non-intrusive observation framework that removes the probe effect.
It reconstructs the full ROS~2 communication state from the domain's discovery packets at no bandwidth cost, then drives an in-kernel filter restricted to the topics the user asks for, lifting only those packets at minimal cost and observing the messages the real subscriber receives.
Its interfaces and recordings match the standard ROS~2 tools.
Across three hardware platforms (laptop, Jetson, and Raspberry Pi), two DDS implementations, and seven robot-operation workloads, ros2probe holds the discovery graph within 0.5\% of an unobserved system, whereas domain-joining tools inflate discovery up to 2.6$\times$ and drop 38.5\% of the subscriber's messages at saturation while ros2probe drops none.
It reports loss with a recall of 1.0, cuts observer CPU and memory by up to 7$\times$ and 28$\times$, and stays practical on the embedded robots where existing tools overload the system.
\end{abstract}



\maketitle

\section{Introduction}\label{section1}

\looseness=-1 Robot Operating System 2~(ROS~2)\cite{macenski2022robot} is an open-source middleware framework for robot software development.
It is the de facto standard system in modern robotics and has been widely adopted in systems ranging from autonomous vehicles and industrial manipulators to service robots.
A robot application consists of multiple nodes with distinct roles, such as sensor processing, motion control, and perception and planning algorithms.
ROS~2 connects these nodes through a layered runtime. An execution model schedules their callbacks, and a communication stack carries their messages from the client libraries down through the middleware to the operating system's network stack.
ROS~2 adopts the Data Distribution Service~(DDS) as its inter-node communication middleware, and each node is interconnected via DDS-based pub/sub communication.
This distributed structure provides modularity and scalability, but no single node can observe the operation of the entire system.
In developing a robot application, monitoring and logging this inter-node communication are essential, whether to evaluate robustness, reproduce and debug failures, or track runtime behavior such as message latency, publication-rate variation, and discovery.
This requires observing the inter-node communication itself, from outside the nodes and in real time.

\looseness=-1 Across the ROS~2 ecosystem, the only way a tool observes messages is to join the domain and subscribe to the target topic.
Every existing tool therefore joins the DDS domain, matches the publisher through discovery, and subscribes to each message directly.
The act of observation thereby enrolls the tool into the publisher's transmission set and discovery graph and perturbs the very system it measures.
This is not a defect of any particular tool but a consequence of the DDS publish/subscribe design.
Much like the observer effect in quantum mechanics, where measuring a system disturbs it, we call this the probe effect~\cite{gait1986probe, mizrahi2024observer}.

\looseness=-1 The probe effect has four distinct manifestations.
First, joining as a participant inflates the discovery plane, adding discovery and control traffic that the unobserved system never carried, even before any topic is observed.
Second, a subscribing observer deserializes every message it receives through the DDS stack and maintains per-topic QoS state, so its CPU and memory grow with the message rate and can overload constrained platforms.
Third, because the observer receives its own separate copy, it loses a different set of messages than the subscriber, so the losses and delays it reports do not match what the subscriber actually experienced.
Fourth, that separate copy is an extra unicast stream, so near link saturation it displaces the original subscriber's messages and becomes real loss on its path.

\looseness=-1 No existing tool resolves the probe effect that arises in robot middleware.
Standard ROS~2 tools (ros2cli, rosbag2) subscribe as ROS~2 nodes and incur all four manifestations.
The monitoring tools built into some DDS implementations avoid the ROS~2 stack overhead but still join the domain as a participant, so they inherit the same four and additionally bind observation to one vendor's implementation, breaking the implementation independence ROS~2 is built on.
The remaining option is to capture packets off the wire from outside the domain, as Wireshark and bpftrace do, which escapes the probe effect but carries a serious limitation of its own.
It exposes only raw packets and cannot reconstruct ROS~2-level state such as the topic graph, per-topic rate, and loss in real time, and lacking a way to select packets it must lift all traffic to user space, so its cost scales with the system's total traffic rather than with what is observed, which is impractical on embedded robots.
What is needed, then, is a tool that meets four requirements at once: no probe effect on the observed system, independence from the DDS implementation, real-time reconstruction of ROS~2 semantic state, and observation cost held to a minimum.

\looseness=-1 This paper presents ros2probe, a non-intrusive observation framework that removes the probe effect by observing ROS~2 entirely from outside the DDS domain.
This paper makes the following contributions.
\begin{enumerate}[leftmargin=*,labelsep=0.4em,itemsep=1pt,parsep=0pt,topsep=2pt]
  \item \textbf{Defining the probe effect.} We formulate the probe effect as a structural consequence of the DDS publish/subscribe design rather than a limitation of individual tools, and decompose it into four measurable manifestations, discovery-plane perturbation, observer resource cost, observation fidelity, and the data-plane traffic and the loss it induces.
  \item \textbf{ros2probe.} We design ros2probe to exploit the fact that DDS carries discovery, topic, and payload information in plaintext over the standard RTPS~\cite{omg2014rtps} wire protocol regardless of vendor. By passively capturing and interpreting RTPS at a network interface it reconstructs the full ROS~2 topic graph and per-topic metrics without creating any endpoint, removing all four manifestations of the probe effect and staying independent of the DDS implementation. The reconstructed discovery in turn drives an in-kernel filter that copies only the topics under observation and follows them as they change at runtime, so its cost scales with the observed topics rather than the total traffic.
  \item \textbf{Extensive comparative evaluation.} We evaluate ros2probe against existing tools across seven robot-operation workloads, two DDS implementations, and three hardware platforms down to embedded boards, through four experiments. First, on discovery transparency, ros2probe holds the discovery graph within 0.5\% of an unobserved system, whereas domain-joining tools inflate it up to 2.6$\times$. Second, on resource overhead, ros2probe cuts observer CPU and memory by up to 7$\times$ and 28$\times$ and runs on embedded boards where existing tools overload the system. Third, on observation fidelity, ros2probe observes losses in exact agreement with the subscriber's actual losses, whereas existing observers report a statistically independent loss set. Fourth, under load, ros2probe succeeds at observation in every scenario the link bandwidth permits, whereas an existing tool inflicts fatal side effects on the real subscriber, whose original transmissions are lost under bandwidth pressure and, with reliable transport, fail to be delivered under retransmission storms.
  \item \textbf{Open-source release.} We release ros2probe as a complete, deployable open-source tool\footnote{Project page: \url{https://csi-dgist.github.io/ros2probe-page/}\\ Source code: \url{https://github.com/csi-dgist/ros2probe}}, whose interfaces and recording formats match the standard ROS~2 tools so it drops into existing workflows unchanged.
\end{enumerate}

\looseness=-1 The remainder of this paper is structured as follows.
Section~\ref{section2} reviews related work across the ROS~2, middleware, and network layers.
Section~\ref{section3} formalizes the probe effect that DDS-domain tools cannot escape and the aggregation gap of network-layer tools, then derives the requirements a faithful observer must meet.
Section~\ref{section4} details the design and implementation of ros2probe, Section~\ref{section5} evaluates it across multiple platforms, DDS implementations, and workloads, and Section~\ref{section6} concludes.

\section{Related Work}\label{section2}
\subsection{ROS~2 Middleware Ecosystem}\label{subsec:ecosystem}

\looseness=-1 As a communication middleware, ROS~2 sits between application code and the operating system's network stack, giving nodes a uniform publish/subscribe interface so they exchange messages without handling transport, discovery, or serialization themselves.
ROS~2 realizes this interface through DDS, and DDS implementations from various vendors exist.
Representative implementations include eProsima's FastDDS~\cite{fastdds}, which ROS~2 adopts by default. The Eclipse Foundation's CycloneDDS~\cite{cyclonedds} aims for small code size and low resource usage, and RTI's commercial Connext DDS~\cite{rticonnext} features rich QoS policies and analytics tools.
Although they differ in vendor, internal structure, and default configuration, they all use RTPS over UDP as the wire protocol.
Meanwhile, ROS~2 also supports transports that do not use RTPS.
Zenoh~\cite{corsaro2023zenoh} is a middleware designed to reduce the burden arising from distributed discovery in DDS, and it uses its own wire protocol different from RTPS.
In this paper, ros2probe targets DDS implementations that use RTPS over UDP as the wire protocol.

\looseness=-1 Despite the existence of multiple implementations, ROS~2 systems can communicate identically regardless of the selected vendor, owing to two mechanisms at different levels.
At the upper level, the ROS Middleware Interface~(RMW) abstracts communication functions into vendor-independent APIs, and each DDS vendor provides an implementation that links these to its own API.
This abstraction allows the same application code to run seamlessly on any DDS implementation.
At the lower level, the RTPS wire protocol defines the packet formats actually exchanged between nodes.
RTPS defines distributed discovery without a central master.
The Simple Participant Discovery Protocol~(SPDP) handles discovery via multicast at the participant level, while the Simple Endpoint Discovery Protocol~(SEDP) handles discovery via unicast at the endpoint level.
During this process, each participant exposes node, topic, type, and QoS information on the wire~\cite{lee2025dependency}, while user data is serialized into CDR and transmitted.

\subsection{Application-Layer Observation Tools}\label{subsec:ros2-tools}

\looseness=-1 Observation tools operating on the ROS~2 stack are broadly divided into two categories.
One acts as a ROS~2 client to observe inter-node communication, while the other embeds instrumentation into ROS~2 to analyze callback execution timing.
Communication-state observation tools subscribe to topics to monitor publication rate, message content, and delivery state.
Topic commands of ros2cli (e.g., \texttt{ros2 topic hz} and \texttt{ros2 topic echo}) output this information in real time, while rosbag2 records messages to SQLite3 and MCAP files for playback.
Execution timing analysis tools instrument the ROS~2 stack to collect execution events at the callback level.
ros2_tracing~\cite{bedard2022ros2_tracing} is an instrumentation framework based on LTTng that records callback, message, and executor events.
CARET~\cite{kuboichi2022caret} analyzes the end-to-end latency of the callback chain on top of it.
The recently proposed eBPF-based tracer~\cite{abaza2024trace} synthesizes the callback dependency graph and execution time using eBPF. 
However, since they focus on internal node execution timing, they are not direct comparison targets for this paper, which addresses inter-node communication state such as message delivery, loss, and retransmission.

\subsection{Middleware-Layer Observation Tools}\label{subsec:mw-tools}

\looseness=-1 Some DDS vendors embed monitoring functions directly within their own implementations.
These middleware-layer tools access the internal state of the DDS stack to provide more detailed statistics and topology information than ROS~2 application tools.
However, enabling these features requires recompilation or that vendor's specific distribution.
FastDDS provides the Statistics Module, which publishes internal statistics such as latency and throughput as topics, and the DDS Recorder, which records arbitrary-type messages to MCAP files.
CycloneDDS's Cyclone DDS Insight displays the node and topic graphs and the speed of data flow between nodes via a GUI, and monitors topics by directly subscribing to their data through a Listener.
RTI Connext visualizes topology and QoS through the Admin Console and aggregates metrics and logs into a dashboard via the Observability Framework.
While these capabilities are powerful, they do not escape the fundamental limitation of the application tools.
Each still participates in the domain, subscribing to the observed topics or requiring extra instrumentation to be enabled inside the participants, so it perturbs the observed system and incurs the probe effect just as a ROS~2 subscriber does.
They are additionally bound to a single vendor's DDS implementation, which prevents their uniform application across ROS~2 environments where multiple vendor implementations coexist.

\subsection{Network-Layer Observation Tools}\label{subsec:net-tools}

\looseness=-1 Network-layer observation tools do not participate in the DDS stack.
Instead, they passively capture packets off the wire.
Wireshark~\cite{wireshark} is a widely used network protocol analyzer that captures packets via libpcap~\cite{mccanne1993bsd} and decodes each packet through its built-in RTPS dissector, displaying fields such as submessage type, sequence number, and endpoint identifier.
With display filters and the tshark CLI, it excels at inspecting individual packet exchanges in detail.
bpftrace~\cite{bpftrace} is a high-level tracing frontend for eBPF~\cite{ebpf2014kernel} that attaches probes to kernel functions and tracepoints to count packet events or analyze latency through the network stack as histograms.
However, both tools are general-purpose and not aware of ROS~2, which limits them in two ways.
First, they expose individual packets and events but cannot interpret their meaning, so they do not aggregate ROS~2 application state such as the topic graph or per-topic delivery in real time.
Second, they have no way to select the relevant traffic by topic as it changes at runtime, so they must capture and process all packets, and their cost grows with the total traffic rather than with what is observed, which is impractical on embedded platforms.

\looseness=-1 ros2probe also observes at this network layer, and like bpftrace it builds on eBPF. Middleware systems have adopted eBPF widely, but for roles different from the one ros2probe needs.
Prior systems place eBPF on the execution path to trace events, as in CaT~\cite{esteves2021cat} and the ROS~2 timing-model tracer of Abaza et al.~\cite{abaza2024trace}, or use it as an active data plane that rewrites and redirects traffic, as in EdgeConnector~\cite{cui2025edgeconnector}.
ros2probe does neither.
It uses eBPF purely for passive, selective capture, and thereby escapes, together, the limitations surveyed across Sections~\ref{subsec:ros2-tools} to~\ref{subsec:net-tools}.
It observes without the probe effect, independently of the DDS implementation, reconstructs ROS~2 state in real time, and keeps its cost to the topics under observation.
The following sections develop this approach.

\section{Background and Motivation}\label{section3}

\subsection{Probe Effect in DDS-Based Tools}\label{subsec:probe-effect}

\looseness=-1 Observation tools that participate in a DDS domain cannot avoid the probe effect, regardless of their implementation, because there is no way to receive data without participating.
To receive data, a DataReader must be created and then discovered and matched by the publisher, after which the publisher treats it as both a delivery target and a QoS negotiation partner.
The rule that only DataReaders matched by discovery can receive data is stipulated by the RTPS standard and shared by all conformant implementations, so no implementation can bypass it.

\begin{figure}[tbp]
  \centering
  \begin{subfigure}[t]{0.49\columnwidth}
    \centering
    \begin{tikzpicture}[
      font=\footnotesize,
      box/.style={draw, rounded corners, minimum width=13mm, minimum height=6mm, align=center, font=\scriptsize},
      rbox/.style={draw=rpred, text=rpred, rounded corners, minimum width=13mm, minimum height=6mm, align=center, font=\scriptsize},
    ]
      \useasboundingbox (-0.7,1.3) rectangle (3.4,-2.8);
      \node[box] (pub) at (0,0) {Publisher};
      \node[box] (s1) at (2.7,0.9) {Subscriber};
      \node[box] (s2) at (2.7,0) {Subscriber};
      \node[rbox] (ob) at (2.7,-1.0) {Observer};
      \draw[->,>=Latex] (pub) -- (s1);
      \draw[->,>=Latex] (pub) -- (s2);
      \draw[->,>=Latex,rpred,thick] (pub) -- node[below,pos=0.65,font=\scriptsize]{$+sf$} (ob);
      \node[draw=rpred, fill=rpred!10, text=rpred, rounded corners, font=\scriptsize, inner sep=2pt] at (1.35,-1.9) {extra copy $\Rightarrow$ probe effect};
    \end{tikzpicture}
    \caption{DDS observer (DataReader)}
    \label{fig:probe-dds}
  \end{subfigure}
  \hfill
  \begin{subfigure}[t]{0.49\columnwidth}
    \centering
    \begin{tikzpicture}[
      font=\footnotesize,
      box/.style={draw, rounded corners, minimum width=13mm, minimum height=6mm, align=center, font=\scriptsize},
      obs/.style={draw, dashed, rounded corners, minimum width=13mm, minimum height=6mm, align=center, font=\scriptsize},
    ]
      \useasboundingbox (-0.7,1.3) rectangle (3.4,-2.8);
      \node[box] (pub) at (0,0) {Publisher};
      \node[box] (s1) at (2.7,0.55) {Subscriber};
      \node[box] (s2) at (2.7,-0.55) {Subscriber};
      \draw[->,>=Latex] (pub) -- (s1);
      \draw[->,>=Latex] (pub) -- (s2);
      \node[obs] (rp) at (1.35,-1.5) {ros2probe};
      \draw[->,>=Latex,dashed] (1.35,-0.275) -- node[right,font=\scriptsize]{copy} (rp);
      \node[draw=rpblue, fill=rpblue!12, text=rpblue, rounded corners, font=\scriptsize, inner sep=2pt] at (1.35,-2.3) {$\Delta B = 0$ (none)};
    \end{tikzpicture}
    \caption{ros2probe (passive capture)}
    \label{fig:probe-probe}
  \end{subfigure}
  \caption{The probe effect on incoming bandwidth. (a) A DDS observer is matched as an extra DataReader, so the publisher sends one more unicast copy ($\Delta B = sf$). (b) ros2probe captures passively and is never matched ($\Delta B = 0$).}
  \label{fig:probe}
  \Description{Two schematic diagrams contrasting observation models: a DDS observer that the publisher matches as an extra DataReader and serves with one additional unicast copy, versus ros2probe, which captures packets passively off the wire and adds no extra traffic.}
\vspace{-1em}
\end{figure}
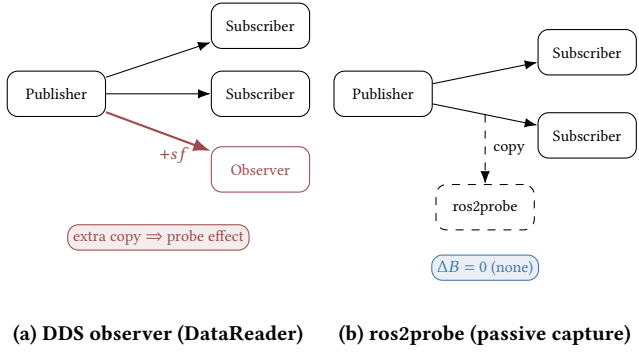

\looseness=-1 The data-plane manifestation of this effect, the growth in incoming bandwidth, can be made precise.
Consider a topic whose publisher emits samples of serialized size $s$ at rate $f$.
Because DDS delivers user data by unicast, the publisher sends one copy to each matched DataReader, so the data-plane bandwidth it produces for $n$ subscribers is
\begin{equation}
B(n) = n \cdot s \cdot f .
\end{equation}
An observation tool that subscribes appears to the publisher as one more DataReader, raising the count from $n$ to $n+1$ and injecting
\begin{equation}
\Delta B = B(n+1) - B(n) = s \cdot f
\end{equation}
of additional traffic, independent of how many subscribers already exist.
Under RELIABLE QoS the publisher must also service the new reader's Heartbeat and AckNack exchange, so the true increase is $\Delta B \ge s \cdot f$.
A DDS-domain observer therefore can never be read-only on the wire. It costs the publisher at least one full unicast stream, and this extra $\Delta B$ is most damaging near link saturation, as Figure~\ref{fig:probe} illustrates.
When the aggregate bandwidth already approaches the link capacity $C$, it has nowhere to go and eats into the delivery of the original subscribers, turning the probe effect into measurable loss on their path, as Section~\ref{subsec:eval-probe} confirms.

\begin{table*}[tbp]
\centering
\caption{Capability comparison of existing ROS~2 observation tools (\fullsupport: supported, \nosupport: not supported).}
\label{tab:tool-comparison}
\begin{tabular}{lcccccccc}
\toprule
Criterion & ros2cli & rosbag2 & FastDDS Stats. & DDS Recorder & Cyclone Insight & Wireshark & bpftrace & \textbf{ros2probe} \\
\midrule
Graph observation   & \fullsupport & \nosupport & \nosupport & \nosupport & \fullsupport & \nosupport & \nosupport & \fullsupport \\
State tracking      & \fullsupport & \nosupport & \fullsupport & \nosupport & \fullsupport & \nosupport & \nosupport & \fullsupport \\
Data recording      & \fullsupport & \fullsupport & \nosupport & \fullsupport & \nosupport & \fullsupport & \nosupport & \fullsupport \\
Middleware-neutral  & \fullsupport & \fullsupport & \nosupport & \nosupport & \nosupport & \fullsupport & \fullsupport & \fullsupport \\
Selectivity         & \fullsupport & \fullsupport & \fullsupport & \fullsupport & \fullsupport & \nosupport & \nosupport & \fullsupport \\
No probe effect     & \nosupport & \nosupport & \nosupport & \nosupport & \nosupport & \fullsupport & \fullsupport & \fullsupport \\
\bottomrule
\end{tabular}
\vspace{-1em}
\end{table*}

\looseness=-1 This limitation cannot be removed as long as observation remains within the DDS domain.
Eliminating it requires observing from outside the domain, where a passive observer creates no DataReader, leaves the publisher's transmission set unchanged, and thus contributes $\Delta B = 0$.

\subsection{Aggregation Gap in Network-Layer Tools}\label{subsec:aggregation-gap}

\looseness=-1 Network-layer tools avoid the probe effect, but no existing one aggregates ROS~2 topology and delivery state in real time, and supplying that aggregation runs into structural constraints whether it is placed in user space or in the kernel.
Wireshark's RTPS dissector interprets each packet independently and statelessly, so it can perform neither topic-graph reconstruction, which correlates packets, nor drop detection based on sequence-number gaps.
To obtain this semantic information, stateful processing is required to track SPDP and SEDP discovery state and accumulate per-topic statistics across the entire packet stream.
This effectively re-implements an observation tool on top of the capture, and because it analyzes capture files offline, it does not operate in real time on a running system.
One might instead perform this processing in the kernel.
However, general-purpose tracing tools such as bpftrace are specialized for aggregating kernel events into maps as counts and histograms, which makes them unsuitable for RTPS parsing and state aggregation.
Even with a custom eBPF program, the complete parsing and aggregation that interprets every submessage and accumulates per-topic state is difficult to perform in the kernel, owing to the verifier's bounded-loop and program-complexity constraints~\cite{vieira2020fast}.
Capture-tool filters must also be specified statically before observation begins and cannot be changed at runtime.
Whether it is a Wireshark capture filter or a bpftrace script, changing the filter requires stopping and restarting the tool, which does not fit a live system in which topics appear and disappear dynamically.
Without a filter, all irrelevant packets are forwarded to user space, so the processing load grows in proportion to the total traffic volume, which is impractical on resource-constrained embedded platforms.
To fill this gap, network-layer capture must be combined with RTPS state tracking and aggregation.

\looseness=-1 The four requirements set out in Section~\ref{section1} (no probe effect, DDS-implementation independence, real-time semantic information, and low overhead) expand into the six concrete, observable criteria of Table~\ref{tab:tool-comparison}, against which we summarize the existing tools examined in Sections~\ref{section2} and~\ref{section3}.
The first three criteria concern the capabilities a tool provides.
Graph observation indicates whether a tool exposes the configuration of running nodes and topics. State tracking indicates whether it computes per-topic publication rate, bandwidth, and drops. Data recording indicates whether it saves messages to files and replays them.
The remaining three criteria concern properties of the observation method.
Middleware neutrality indicates whether it is independent of any specific DDS vendor. Selectivity indicates whether the tool can restrict observation to the topics of interest so that its cost scales with the observed traffic rather than the total. No probe effect indicates whether observation leaves the target's behavior and traffic unchanged.
No existing tool satisfies all four at once, and this gap is the starting point for the design of ros2probe.

\section{Design and Implementation}\label{section4}

\subsection{Pipeline Overview}\label{subsec:pipeline}

\begin{figure*}[tbp]
  \centering
  \includegraphics[width=\textwidth]{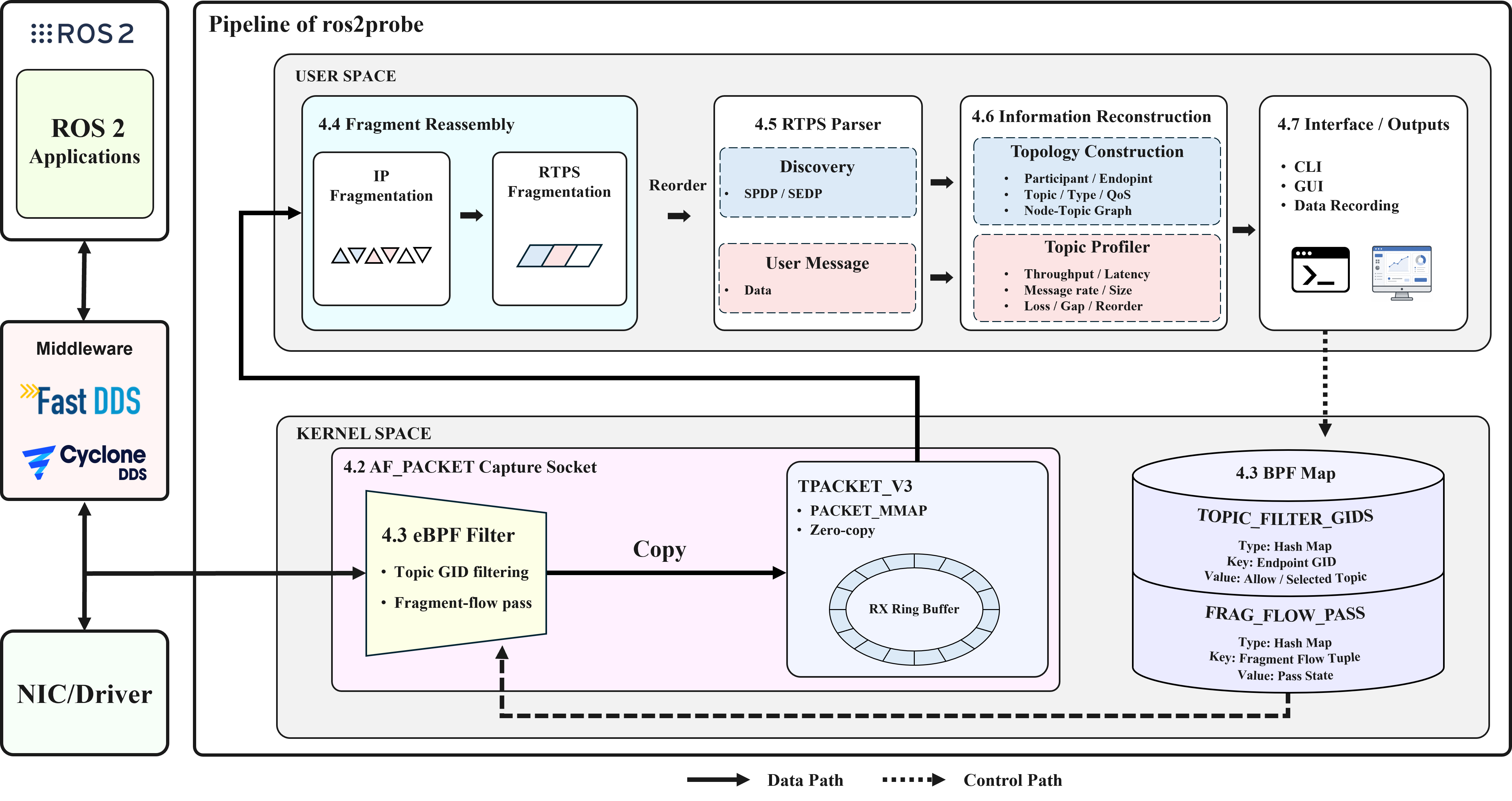}
  \caption{Pipeline of ros2probe.}
  \label{fig:pipeline}
  \Description{Block diagram of the ros2probe pipeline, showing a kernel-space packet-selection layer feeding a user-space stage that reassembles packets and splits them into a discovery path that builds the topic graph and a data path that computes per-topic metrics.}
\vspace{-1em}
\end{figure*}

\looseness=-1 As shown in Figure~\ref{fig:pipeline}, ros2probe, implemented in Rust, consists of a packet-selection layer in kernel space and a parsing-and-analysis layer in user space.
The kernel layer passively receives copies of packets at the network interfaces.
Because this observation merely reads copies at the network layer, outside the ROS~2 application, and does not intervene in the application itself, it makes no modification to the nodes of the observed system.
Because the observation tool does not participate in the DDS domain, publishers transmit no additional data.
Packets unrelated to the topics of interest are not passed up to user space, which confines the parsing cost to the load of the topics under observation.
The user-space layer comprises three stages.
A reassembly stage first restores fragmented packets into their original messages.
Each restored message is then branched onto one of two paths according to its role.
The discovery path parses the inter-node negotiation messages and maintains the running ROS~2 topic graph (topic names, types, and QoS) in real time.
Whenever a new topic is discovered, it reflects the result into the kernel filter to update the scope of selection.
The data path extracts per-message identifiers and timestamps and recovers the originating node name.
The Topic Profiler derives the per-topic publication rate, bandwidth, and loss from the data path's measurements, and, given synchronized clocks, latency as well.
ros2probe exposes these, together with the topic graph from the discovery path, through a CLI, a GUI, and files.

\subsection{Packet Capture}\label{subsec:capture}
To avoid perturbing the target system, the observation must receive packets from outside the DDS domain.
To this end, ros2probe uses Linux's AF_PACKET socket.
AF_PACKET is a socket family that delivers copies of the frames traversing a network interface directly to user space, bypassing the host protocol stack, and thus allows RTPS traffic to be received without creating a DDS DataReader.
ros2probe opens one socket on every network interface of the observation host, capturing copies not only of the RTPS frames crossing the network segment but also of topics published locally over loopback on the same host.
Because these sockets only read copies and transmit no packets, ros2probe is registered in the DDS domain as neither a subscriber nor a participant, and it does not interfere with the delivery of the original traffic in any way.
To lower the receive cost, ros2probe uses TPACKET_V3, the block-based version of AF_PACKET's memory-mapped ring buffer.
In TPACKET_V3, the kernel packs multiple frames into a shared memory block and hands them to user space a block at a time, so frames are read in bulk without incurring a system call or a copy per packet.
Each frame arrives with a kernel-applied receive timestamp, which ros2probe uses as the message arrival time.

\looseness=-1 This capture model also delimits what ros2probe can observe across DDS transports.
Discovery is always visible. SPDP and SEDP are exchanged over UDP even when participants enable the shared-memory~(SHM) transport for user data, so the topic graph is reconstructed regardless of the transport in use.
User data sent over SHM, however, never reaches a network interface and is therefore outside ros2probe's view.
This is not a limitation specific to ros2probe but a structural property of passive wire observation. Shared-memory traffic leaves no trace on the wire, so no passive observer can capture it without intruding on the participants.
Non-intrusive tracking or recording of SHM-only user data is thus fundamentally impossible at the wire level.

\looseness=-1 The same boundary applies to data that is never transmitted at all.
A topic with no matched subscriber produces no wire traffic, even if a sensor driver keeps calling publish, because the middleware sends nothing until a reader is discovered.
ros2probe therefore reports no data on such a topic, which faithfully reflects the absence of any subscriber.
A DDS-domain tool cannot report this state without altering it. By subscribing, ros2cli tools become the first consumers of a topic that nobody was consuming and report it as actively published, perturbing the very condition they measure.

\subsection{Kernel-Selective Filtering}\label{subsec:filter}
Passing every received packet up to user space makes the processing cost proportional to the total traffic, which is hard to sustain in embedded settings.
ros2probe therefore restricts the packets it processes to the topics of interest, and an eBPF program attached to the AF_PACKET socket performs this selection.
This eBPF program is written with Aya~\cite{aya}, an eBPF framework for Rust, to match the Rust codebase.
The filter runs in kernel space before a frame is placed on the RX ring and decides whether to pass the frame up to user space, copying it to the capture socket, or to ignore it.
A frame the filter ignores is simply never copied, and the original packet's normal delivery is unaffected either way.

\begin{figure}[tbp]
  \centering
  \begin{tikzpicture}[
    cell/.style={draw, anchor=north west, align=center, font=\scriptsize, inner sep=1.5pt},
  ]
    \node[cell, fill=black!8, minimum width=1.6cm, minimum height=0.55cm] at (0,0) {Ethernet\\(14\,B)};
    \node[cell, fill=black!8, minimum width=1.9cm, minimum height=0.55cm] at (1.6,0) {IPv4\\(20\,B)};
    \node[cell, fill=black!8, minimum width=1.0cm, minimum height=0.55cm] at (3.5,0) {UDP\\(8\,B)};
    \node[cell, fill=rpblue!18, minimum width=1.5cm, minimum height=0.55cm] at (4.5,0) {UDP\\payload};

    \draw[dashed] (4.5,-0.55) -- (0,-1.15);
    \draw[dashed] (6.0,-0.55) -- (6.0,-1.15);

    \node[cell, fill=rpblue!18, minimum width=6cm, minimum height=0.55cm] at (0,-1.15) {\textbf{RTPS message} (UDP payload)};

    \node[cell, fill=rpblue!8, minimum width=6cm, minimum height=0.55cm] at (0,-1.70) {\texttt{RTPS} magic (4\,B)};
    \node[cell, fill=rpblue!8, minimum width=3cm, minimum height=0.55cm] at (0,-2.25) {version (2\,B)};
    \node[cell, fill=rpblue!8, minimum width=3cm, minimum height=0.55cm] at (3,-2.25) {vendorId (2\,B)};
    \node[cell, fill=rpblue!35, minimum width=6cm, minimum height=1.65cm] at (0,-2.80) {GUID prefix (12\,B)};

    \node[cell, fill=black!8, minimum width=1.5cm, minimum height=0.55cm] at (0,-4.45) {subId\\(1\,B)};
    \node[cell, fill=black!8, minimum width=1.5cm, minimum height=0.55cm] at (1.5,-4.45) {flags\\(1\,B)};
    \node[cell, fill=black!8, minimum width=3cm, minimum height=0.55cm] at (3,-4.45) {octetsToNext\\Header (2\,B)};
    \node[cell, fill=black!8, minimum width=3cm, minimum height=0.55cm] at (0,-5.00) {extraFlags (2\,B)};
    \node[cell, fill=black!8, minimum width=3cm, minimum height=0.55cm] at (3,-5.00) {octetsTo\\InlineQos (2\,B)};
    \node[cell, fill=black!8, minimum width=6cm, minimum height=0.55cm] at (0,-5.55) {reader EntityId (4\,B)};
    \node[cell, fill=rpred!30, minimum width=6cm, minimum height=0.55cm] at (0,-6.10) {writer EntityId (4\,B)};
    \node[cell, fill=black!8, minimum width=6cm, minimum height=1.10cm] at (0,-6.65) {writerSN (8\,B)};
    \node[cell, dashed, minimum width=6cm, minimum height=0.55cm] at (0,-7.75) {inlineQoS (optional)};
    \node[cell, minimum width=6cm, minimum height=0.55cm] at (0,-8.30) {serializedPayload (CDR)\,\dots};

    \draw[semithick] (-0.06,-1.75) -- (-0.18,-1.75) -- (-0.18,-4.35) -- (-0.06,-4.35);
    \draw[semithick] (-0.06,-4.55) -- (-0.18,-4.55) -- (-0.18,-8.80) -- (-0.06,-8.80);
    \node[rotate=90, anchor=center, font=\scriptsize] at (-0.48,-3.05) {RTPS header (20\,B)};
    \node[rotate=90, anchor=center, font=\scriptsize] at (-0.48,-6.675) {DATA submessage};

    \node[cell, fill=rpblue!35, minimum width=4mm, minimum height=4mm] at (0,-9.25) {};
    \node[anchor=west, font=\scriptsize] at (0.35,-9.45) {GUID prefix (12\,B)};
    \node[cell, fill=rpred!30, minimum width=4mm, minimum height=4mm] at (3.0,-9.25) {};
    \node[anchor=west, font=\scriptsize] at (3.35,-9.45) {writer EntityId (4\,B)};
    \node[anchor=west, font=\scriptsize] at (0,-10.0) {$\Rightarrow$ writer GID (16\,B) $=$ GUID prefix $+$ writer EntityId};
  \end{tikzpicture}
  \caption{RTPS DATA submessage layout and the writer GID (\texttt{inlineQoS} is optional). The filter builds the 16-byte writer GID from the 12-byte GUID prefix and the 4-byte writer EntityId, then looks it up in \texttt{TOPIC\_FILTER\_GIDS}. DATAFRAG inserts fragment fields after writerSN but forms the same writer GID.}
  \label{fig:rtps}
  \Description{Diagram of an RTPS DATA submessage layout and the assembly of the 16-byte writer GID from the 12-byte GUID prefix and the 4-byte writer EntityId.}
\vspace{-1em}
\end{figure}
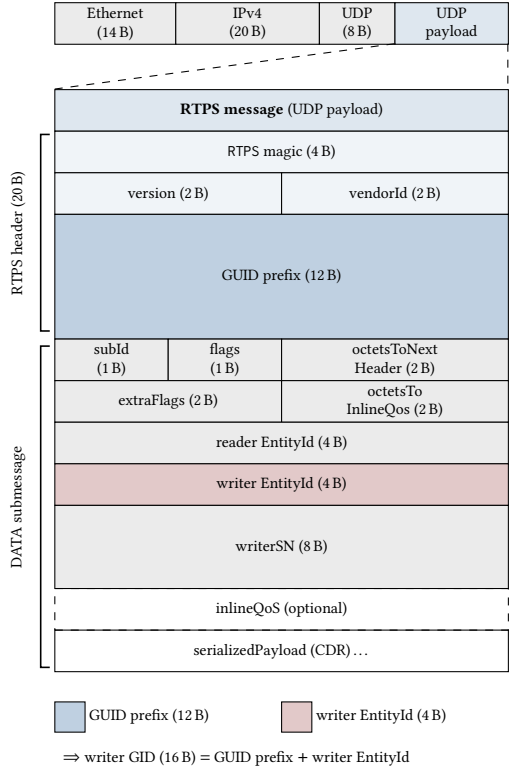

\looseness=-1 The frames the filter sees carry an RTPS message after the Ethernet, IP, and UDP headers.
An RTPS message begins with a 20-byte header. The first 4 bytes are the signature \texttt{RTPS}, followed by the protocol version (2 bytes), the vendor ID (2 bytes), and a 12-byte GUID prefix.
The GUID prefix identifies the DDS participant that sent the packet.
The header is followed by one or more submessages, each delimited by a 4-byte header carrying its kind, flags, and length.
Among them, the DATA and DATAFRAG submessages that carry user data contain a 4-byte entity ID for each of the destination and source endpoints, together with the message sequence number.
As Figure~\ref{fig:rtps} shows, the 16-byte GID that globally identifies an endpoint is formed by appending this entity ID (4 bytes) to the GUID prefix (12 bytes) from the header, indicating which endpoint of which participant it refers to.

\looseness=-1 The filter examines this structure from the cheapest checks to the most specific, as Algorithm~\ref{alg:filter} details.
Beyond the L2--L4 and RTPS-signature checks that skip most of the traffic not under observation, two cases drive the decision.
Discovery traffic is always passed without a GID lookup. SPDP and SEDP messages carry the fixed source entity IDs reserved by the RTPS standard (SPDP participant \texttt{0x000100c2}, SEDP publications \texttt{0x000003c2}, SEDP subscriptions \texttt{0x000004c2}), and passing them keeps ros2probe's view of the topic graph current.
For user data, the filter forms the source and destination GIDs from the GUID prefix and the writer and reader entity IDs, and passes a frame only when either GID belongs to the topics of interest.

\looseness=-1 This set of topics of interest changes at runtime as discovery proceeds, and the decision for an IP-fragmented message must be carried across its fragments, so it cannot be expressed by static rules alone.
ros2probe resolves this with two BPF maps shared between the kernel and user space.
The \texttt{TOPIC_FILTER_GIDS} map, which holds the GIDs of the endpoints of topics of interest, is only read by the kernel and is populated by user space according to the discovery results.
The \texttt{FRAG_FLOW_PASS} map records the fragment flows that have been passed.
It exists for the middle and trailing IP fragments that lack an RTPS header, so that when the first fragment is passed the kernel records the flow, each trailing fragment is then handled identically by merely looking up that record, and the entry is removed at the last fragment.
This map is managed as an LRU so that orphan entries, whose tail fragment never arrives, are evicted automatically.
Because the set of topics of interest is updated at runtime in this way, ros2probe adapts to the dynamic creation and removal of topics that static filters cannot handle, without restarting the capture.

\looseness=-1 Among passive-capture mechanisms, ros2probe uses an eBPF socket filter on AF_PACKET because it runs on the observation host with no special network hardware, leaves the original datagrams untouched, and selects through maps that user space updates at runtime~\cite{scholz2018performance}.

\begin{algorithm}[t]
\caption{Per-frame eBPF socket-filter decision.}
\label{alg:filter}
\begin{algorithmic}[1]
\Require frame $f$, shared maps \textsc{TopicGids}, \textsc{FragPass}
\Ensure \textbf{pass} (copy $f$ to the capture socket) or \textbf{ignore}
\If{ethertype$(f) \notin \{\text{IPv4}, \text{IPv6}\}$ \textbf{ or } ipproto$(f) \neq \text{UDP}$}
    \State \Return \textbf{ignore} \Comment{cheap L2--L4 checks}
\EndIf
\If{$f$ is a non-first IP fragment} \Comment{no RTPS header present}
    \State \Return flowkey$(f) \in \textsc{FragPass}$~?~\textbf{pass} : \textbf{ignore}
\EndIf
\If{udppayload$(f)$ does not start with \texttt{"RTPS"}}
    \State \Return \textbf{ignore}
\EndIf
\State $g \gets$ GUID prefix from the RTPS header \Comment{12\,B}
\For{each DATA/DATAFRAG submessage among the first few of $f$}
    \State $w \gets$ writer entity ID; \quad $r \gets$ reader entity ID \Comment{4\,B each}
    \If{$w$ is a reserved discovery ID (SPDP, SEDP-pub, SEDP-sub)}
        \State \Call{RecordIfFragmented}{$f$}; \quad \Return \textbf{pass} \Comment{always pass discovery}
    \EndIf
    \If{$(g \,\Vert\, w) \in \textsc{TopicGids}$ \textbf{ or } $(g \,\Vert\, r) \in \textsc{TopicGids}$}
        \State \Call{RecordIfFragmented}{$f$}; \quad \Return \textbf{pass} \Comment{topic of interest}
    \EndIf
\EndFor
\State \Return \textbf{ignore}
\Statex
\Function{RecordIfFragmented}{$f$}
    \If{$f$ is the first of an IP-fragmented datagram}
        \State insert flowkey$(f)$ into \textsc{FragPass}
    \EndIf
\EndFunction
\end{algorithmic}
\end{algorithm}

\subsection{Fragment Reassembly}\label{subsec:reassembly}
The parsing layer operates on complete RTPS messages, but large messages arrive split on the wire in two stages.
A UDP datagram exceeding the MTU (typically 1,500 B) is divided into several fragments at the IP layer, and a large sample exceeding the IP datagram limit (about 64 KB) is further split by the DDS implementation into a DATAFRAG sequence at the RTPS level.
ros2probe restores the captured packets into complete RTPS messages and forwards them to either the discovery or the data path depending on the packet kind, performing reassembly independently per interface.
An IP-fragmented packet is reassembled by the standard method once all fragments of a flow have arrived, while a non-fragmented packet is passed through as is.
If the resulting RTPS message carries DATAFRAG, its fragments are gathered by the source GID and sequence number and reassembled again.
Otherwise, the message is already complete and is forwarded as is.
The payload itself is not interpreted here, which is left to the discovery and data paths.
Because ros2probe is a passive observer, however, it cannot request retransmission of lost fragments.
It therefore merely caps the number of incomplete flows it tracks and evicts the oldest when the cap is exceeded, so a flow that is never completed is eventually discarded.

\subsection{RTPS Parser}\label{subsec:parser}
The complete RTPS messages reconstructed in Section~\ref{subsec:reassembly} are parsed and split into two streams. The discovery stream feeds the topic-graph reconstruction and the data stream the per-topic metrics, both detailed in Section~\ref{subsec:metrics}.

\looseness=-1 On the data path, for these measurements to be accurate, each sample must be counted exactly once.
ros2probe, however, captures frames bidirectionally on every interface, so it may observe the same sample more than once.
For example, a sample delivered locally over loopback is caught both as the outgoing frame and as the returning incoming frame.
From each reassembled RTPS message, ros2probe extracts the arrival timestamp together with the source GID and sequence number, and at the point where the outputs of the per-interface workers merge into a single loop, it deduplicates by the (source GID, sequence number) pair before handing the message to the Topic Profiler.

\looseness=-1 The key to keeping this path cheap is that the message payload is never deserialized.
A DataReader-based tool must deserialize every received message according to its type and maintain per-topic QoS state, whereas the per-topic publication rate, bandwidth, and loss are derived from the message metadata alone rather than from the payload content.
ros2probe therefore extracts only the source GID, sequence number, size, and arrival time from each DATA message and hands them to the Topic Profiler, without interpreting the CDR payload itself.

\subsection{Information Reconstruction}\label{subsec:metrics}
From the parsed streams, ros2probe reconstructs the two products that its CLI and GUI present, the ROS~2 topic graph and the per-topic metrics.

\looseness=-1 Because ros2probe does not participate in the DDS domain, it cannot query DDS for the running topology.
Instead, it reconstructs the topology directly by parsing discovery traffic.
The data structure that holds this result is the discovery table, which gathers the participants, endpoints, topic names, types, and QoS parsed from SPDP and SEDP according to the RTPS standard, forming the skeleton of the topic graph.
ros2probe manages the registration, update, and removal of entries in this table through discovery messages. An ALIVE message newly registers a participant or endpoint or updates existing information, while DISPOSE and UNREGISTER messages signal disposal and deregistration and remove the entry.
Each entry is given a time-to-live (TTL) so that even if a participant terminates abnormally without a DISPOSE, the entry is cleaned up automatically upon expiration.
When a participant disappears, all of its endpoints are cascade-removed so that no residual information is left in the table.
Which participant an endpoint belongs to is identified by the participant GUID that SEDP carries as a standard parameter, so this works on every RTPS implementation without vendor-specific extensions.
We confirmed this in practice. Under CycloneDDS as well as FastDDS, ros2probe reconstructs the topic graph and per-topic metrics correctly.

\looseness=-1 Even when constructed and managed in this way, the discovery table is only the skeleton of the topic graph and does not capture which ROS~2 node each endpoint belongs to.
This is because DDS discovery (SEDP) exposes only participants, endpoints, topics, and QoS and is unaware of the ROS~2 notion of a "node". A single participant may contain multiple nodes, and which endpoint belongs to which node is not revealed at the DDS level.
ROS~2 fills this gap with \texttt{ros_discovery_info}, a built-in topic published by the RMW layer.
Each message on this topic carries the list of nodes belonging to one participant, and for each node it includes the name, namespace, and the GIDs of the reader and writer endpoints that the node owns.
ros2probe parses the DATA of this topic on the data path and links each endpoint GID to a node name, thereby attaching node identity to the discovery table and completing the topic graph.

\looseness=-1 Topology information (the topic graph and QoS) and real-time measurements (publication rate, bandwidth, and loss) differ fundamentally in their update frequency.
The former changes only when a node starts or stops, whereas the latter is aggregated on every message arrival.
Combining the two states into a single structure would let discovery updates interfere with the measurement path.
ros2probe therefore separates them and treats each independently.
The topology path is maintained continuously, whereas the measurement path is opened only on demand, per topic, when the user requests it.

\looseness=-1 Among the measurements, the publication rate, bandwidth, and loss are obtained from metadata alone, without the message payload.
The Topic Profiler groups the events forwarded from the RTPS parser by topic and computes the publication rate and bandwidth as the number of messages and bytes per unit time, and it detects loss from gaps in the per-writer RTPS sequence numbers.
Because these gaps are computed from the very frames that travel the wire to the subscriber, not from an independent copy sent to an observer endpoint, ros2probe's loss and delay reflect the subscriber's actual delivery path. 
This closes the observation-reliability gap of Section~\ref{subsec:probe-effect}, where a DDS-domain observer's separate unicast stream may drop packets the subscriber never loses, and vice versa.
The loss reported here is defined as the per-writer sequence-number gaps counted within the observation window, which differs from the final delivery-loss rate under RELIABLE QoS, where retransmissions may later recover the missing samples~\cite{park2025analytical, lee2025probabilistic}.
Latency, by contrast, is computed from the std_msgs/Header stamp when a message carries one, assuming the publisher and observation host clocks are synchronized.

\subsection{Interface and Output}\label{subsec:interface}
ros2probe is operated through a command-line client, \texttt{rp}.
The observer is started with \texttt{rp run}, after which it continuously maintains the topic graph from discovery.
Per-topic measurement, in contrast, takes place only on demand.
A measurement session for a topic opens only when the user issues a command such as \texttt{rp topic hz} or \texttt{rp bag record}, at which point the topic's GID is registered in the kernel filter so that only that topic's data is copied to user space.
In the idle state, with no observe command, only discovery is ever passed up to user space and no data topic is copied, so the very cost of passing data up, parsing it, and aggregating it does not arise.
Beyond this command-line client, ros2probe also provides a GUI that renders the live topic graph and per-topic metrics.

\looseness=-1 For ros2probe to serve as an immediate replacement for existing tools, its output and recording formats must match those of the existing tools so that a user can swap in the tool alone.
To this end, for the commands it provides, \texttt{rp} mirrors ros2cli's command structure and output format. It offers introspection over \texttt{topic}, \texttt{node}, \texttt{service}, and \texttt{action}, and recording over \texttt{bag}, as in \texttt{rp topic hz}.
Existing workflows, such as scripts that parse hz output or CI pipelines, can therefore be reused without modification.
Recordings are written in rosbag2's MCAP format and replay directly with \texttt{ros2 bag play}.
However, because synchronous disk I/O contends with packet processing under workloads of several hundred Mbps and induces loss, MCAP writing is offloaded to a dedicated thread.

\section{Evaluation}\label{section5}
\begin{figure}[tbp]
  \centering
  \includegraphics[width=0.7\linewidth]{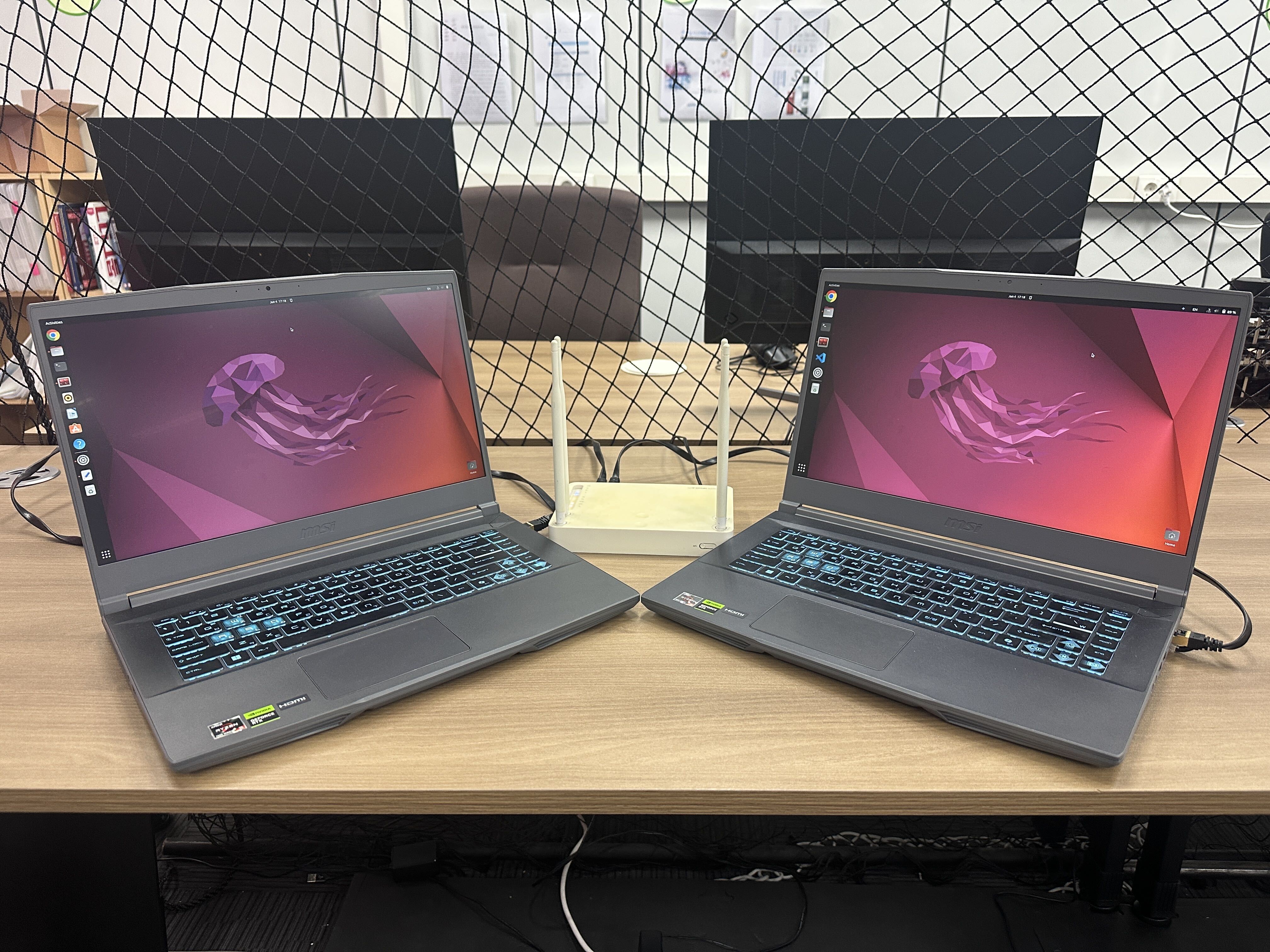}
  \caption{Cross-host setup common to all experiments: a publisher laptop and an identical receiver laptop.}
  \label{fig:testbed}
  \Description{Photograph of the cross-host testbed: a publisher laptop and an identical receiver laptop connected over a wired link.}
\vspace{-1em}
\end{figure}

We design four experiments to verify that ros2probe observes a running ROS~2 system without perturbing it and remains practical even on resource-constrained platforms.
Experiment~1, the discovery-transparency verification of Section~\ref{subsec:eval-discovery}, shows that ros2probe leaves no trace in the DDS discovery graph.
Experiment~2, the resource-overhead evaluation of Section~\ref{subsec:eval-resource}, compares the observer's CPU and memory cost across platforms and shows that ros2probe remains practical even on resource-constrained embedded platforms.
Experiment~3, the observation-fidelity test of Section~\ref{subsec:eval-fidelity}, shows that under packet loss ros2probe records exactly the messages the subscriber received, whereas a DDS-domain observer does not.
Experiment~4, the probe-effect verification of Section~\ref{subsec:eval-probe}, shows that under any workload ros2probe does not affect the incoming traffic or the original subscriber's drop rate.
All experiments run on Ubuntu 22.04 (Linux kernel 5.15) with ROS~2 Humble, and all quantitative results below use FastDDS 2.6.11 as the middleware.
ros2probe is not tied to this implementation. It also runs unmodified on CycloneDDS 0.10.5, where it reconstructs the topic graph and per-topic metrics correctly.
As Figures~\ref{fig:testbed} and~\ref{fig:platforms-photo} show, the two platforms are connected across hosts by a commodity router. All experiments use a wired 1\,Gbps Ethernet link, and the probe-effect experiment of Section~\ref{subsec:eval-probe} additionally sweeps a wireless link.

\begin{table}[tbp]
\centering
\caption{Evaluation platforms.}
\label{tab:platforms}
\footnotesize
\begin{tabular}{llccc}
\toprule
Platform & CPU / SoC & Cores & Clock & RAM \\
\midrule
Laptop & AMD Ryzen 5 7535HS & 6C\,/\,12T & 4.6\,GHz & 16\,GB \\
Jetson Orin NX 16GB & Arm Cortex-A78AE & 8C & 2.0\,GHz & 16\,GB \\
Raspberry Pi 4B & BCM2711 (Cortex-A72) & 4C & 1.5\,GHz & 8\,GB \\
\bottomrule
\end{tabular}
\vspace{-1em}
\end{table}

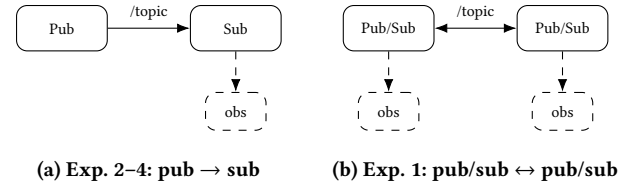
\begin{figure}[tbp]
  \centering
  \begin{subfigure}[t]{0.49\columnwidth}
    \centering
    \begin{tikzpicture}[font=\scriptsize,
      n/.style={draw, rounded corners, minimum width=12mm, minimum height=6mm, align=center, inner sep=1pt},
      o/.style={draw, dashed, rounded corners, minimum width=8mm, minimum height=5mm, align=center, inner sep=1pt}]
      \useasboundingbox (-0.7,0.4) rectangle (3.0,-1.5);
      \node[n] (pub) at (0,0) {Pub};
      \node[n] (sub) at (2.3,0) {Sub};
      \draw[->,>=Latex] (pub) -- node[above]{/topic} (sub);
      \node[o] (obs) at (2.3,-1.1) {obs};
      \draw[->,dashed,>=Latex] (sub) -- (obs);
    \end{tikzpicture}
    \caption{Exp.\ 2--4: pub $\to$ sub}
    \label{fig:topo-13}
  \end{subfigure}
  \hfill
  \begin{subfigure}[t]{0.49\columnwidth}
    \centering
    \begin{tikzpicture}[font=\scriptsize,
      n/.style={draw, rounded corners, minimum width=12mm, minimum height=6mm, align=center, inner sep=1pt},
      o/.style={draw, dashed, rounded corners, minimum width=8mm, minimum height=5mm, align=center, inner sep=1pt}]
      \useasboundingbox (-0.7,0.4) rectangle (3.0,-1.5);
      \node[n] (a) at (0,0) {Pub/Sub};
      \node[n] (b) at (2.3,0) {Pub/Sub};
      \draw[<->,>=Latex] (a) -- node[above]{/topic} (b);
      \node[o] (oa) at (0,-1.1) {obs};
      \node[o] (ob) at (2.3,-1.1) {obs};
      \draw[->,dashed,>=Latex] (a) -- (oa);
      \draw[->,dashed,>=Latex] (b) -- (ob);
    \end{tikzpicture}
    \caption{Exp.\ 1: pub/sub $\leftrightarrow$ pub/sub}
    \label{fig:topo-2}
  \end{subfigure}
  \caption{Workload topology: a one-way publisher-to-subscriber path in Experiments~2--4 versus mutually discovering participants, each a publisher and subscriber, in Experiment~1.}
  \label{fig:disc-topo}
  \Description{Two workload topology diagrams: a one-way publisher-to-subscriber path used in Experiments 2 to 4, and two mutually discovering participants that are each both publisher and subscriber, used in Experiment 1.}
\vspace{-1em}
\end{figure}

\begin{figure*}[tbp]
  \centering
  \includegraphics[width=0.85\textwidth]{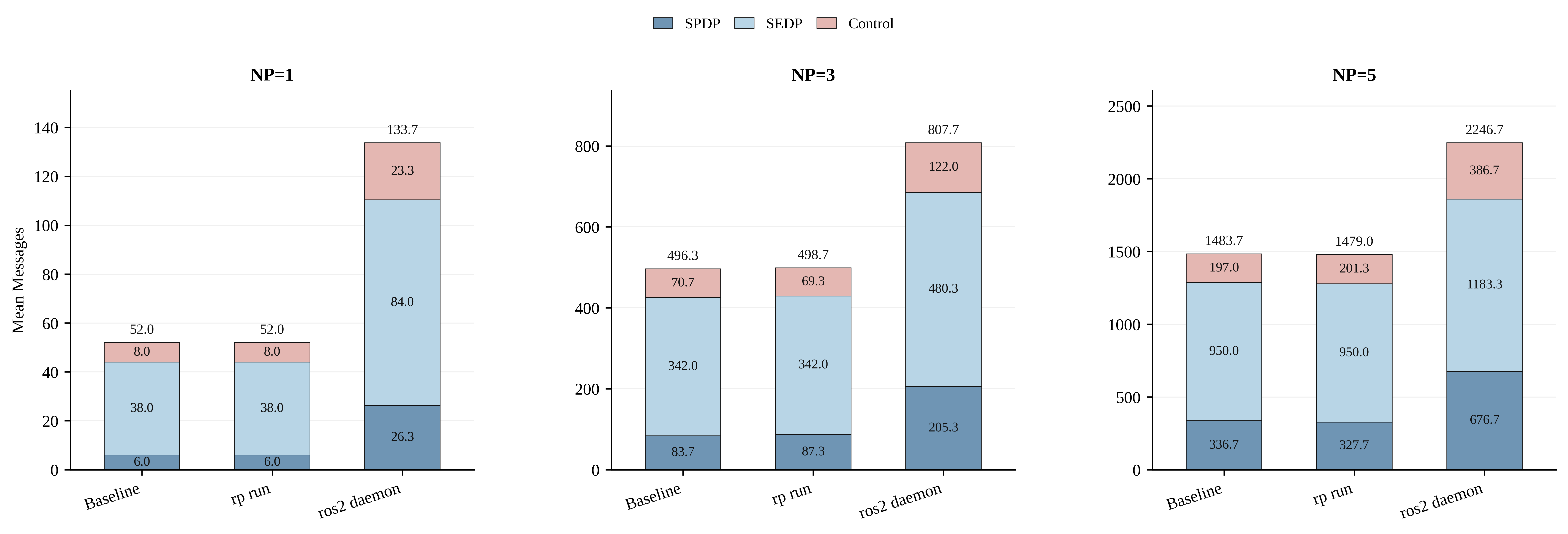}
  \caption{Discovery messages (SPDP, SEDP, Control) under the three conditions at participant counts NP $=$ 1, 3, 5, averaged over 10 runs per condition.}
  \label{fig:exp2}
  \Description{Grouped bar charts of SPDP, SEDP, and Control discovery message counts at participant counts of 1, 3, and 5, showing that rp run matches the baseline while the ros2 daemon inflates all three message types.}
\vspace{-1em}
\end{figure*}

\subsection{Discovery Transparency}\label{subsec:eval-discovery}
ros2probe creates no DDS Participant, so it should leave no trace in the discovery graph.
In contrast, the ros2 daemon that ros2cli commands rely on joins the domain as a DDS Participant to maintain the ROS graph cache.
Figure~\ref{fig:disc-topo} contrasts the two. Unlike the other three experiments, which stream from a single publisher to a single subscriber, this experiment makes every participant both a publisher and a subscriber so that the participants mutually discover, and it sweeps the participant count NP over 1, 3, and 5.
Each participant hosts a fixed two endpoints, one publisher and one subscriber.
Figure~\ref{fig:exp2} reports the measured discovery messages.
The three conditions (baseline, rp run, and ros2 daemon) are each repeated 10 times.

Each run first launches the condition's observer (none for baseline; \texttt{rp run} for rp run; \texttt{ros2 daemon start} for ros2 daemon) and lets it stabilize for 10 seconds, then starts capturing the RTPS discovery traffic (UDP 7400--7600) with tshark on the observation host and immediately launches the participants together.
Stabilizing the observer before the workload separates the tool's own startup traffic from the workload's discovery burst, and the capture starts just before the workload so as not to miss the initial SPDP/SEDP packets.
The captured pcap is interpreted with an RTPS dissector to count SPDP, SEDP, Control, and total discovery packets.

\looseness=-1 rp run is indistinguishable from baseline at every scale. Its total discovery messages stay within about 0.5\% of baseline (52 vs.\ 52 at NP$=$1 and 1484 vs.\ 1484 at NP$=$5), confirming that ros2probe does not touch the discovery graph.
The ros2 daemon, in contrast, inflates discovery at every scale and across all three message types, raising the total by 2.6$\times$, 1.6$\times$, and 1.5$\times$ at NP$=$1, 3, and 5 (52$\rightarrow$134, 496$\rightarrow$808, 1484$\rightarrow$2247).

\looseness=-1 Adding the daemon as a participant inflates all three kinds.
SPDP grows because a new participant periodically multicasts its own liveliness and must be announced to the others.
SEDP grows because endpoint announcements are unicast and must now also be exchanged with the daemon (38$\rightarrow$84, 342$\rightarrow$480, 950$\rightarrow$1183), even though the workload endpoints are unchanged.
Control (Heartbeat/AckNack) grows because the daemon's built-in endpoints require reliable synchronization with every existing participant.
Discovery itself already grows super-linearly with NP (baseline 52$\rightarrow$496$\rightarrow$1484), so although the daemon's relative overhead shrinks at scale, its absolute cost grows ($+82$, $+311$, $+763$ messages).
rp run creates no participant, so it adds nothing at any scale.
Because ros2cli commands operate through this daemon by default, in dense systems merely using ros2cli enlarges discovery traffic, whereas ros2probe does not participate in discovery and causes no such perturbation.

\subsection{Resource Overhead}\label{subsec:eval-resource}

\begin{table}[tbp]
\centering
\caption{Resource-overhead stress workloads.}
\label{tab:stress}
\footnotesize
\begin{tabular*}{\columnwidth}{@{\extracolsep{\fill}}llrrr}
\toprule
Scenario & Topic & Payload & Rate & Est.\ bandwidth \\
\midrule
ST100 & \texttt{/stress} & 64 KiB & 100 Hz & $\sim$52 Mbps \\
ST500 & \texttt{/stress} & 64 KiB & 500 Hz & $\sim$262 Mbps \\
ST1000 & \texttt{/stress} & 64 KiB & 1000 Hz & $\sim$524 Mbps \\
\bottomrule
\end{tabular*}
\vspace{-1em}
\end{table}

\begin{figure}[tbp]
  \centering
  \begin{subfigure}{0.49\linewidth}
    \includegraphics[width=\linewidth]{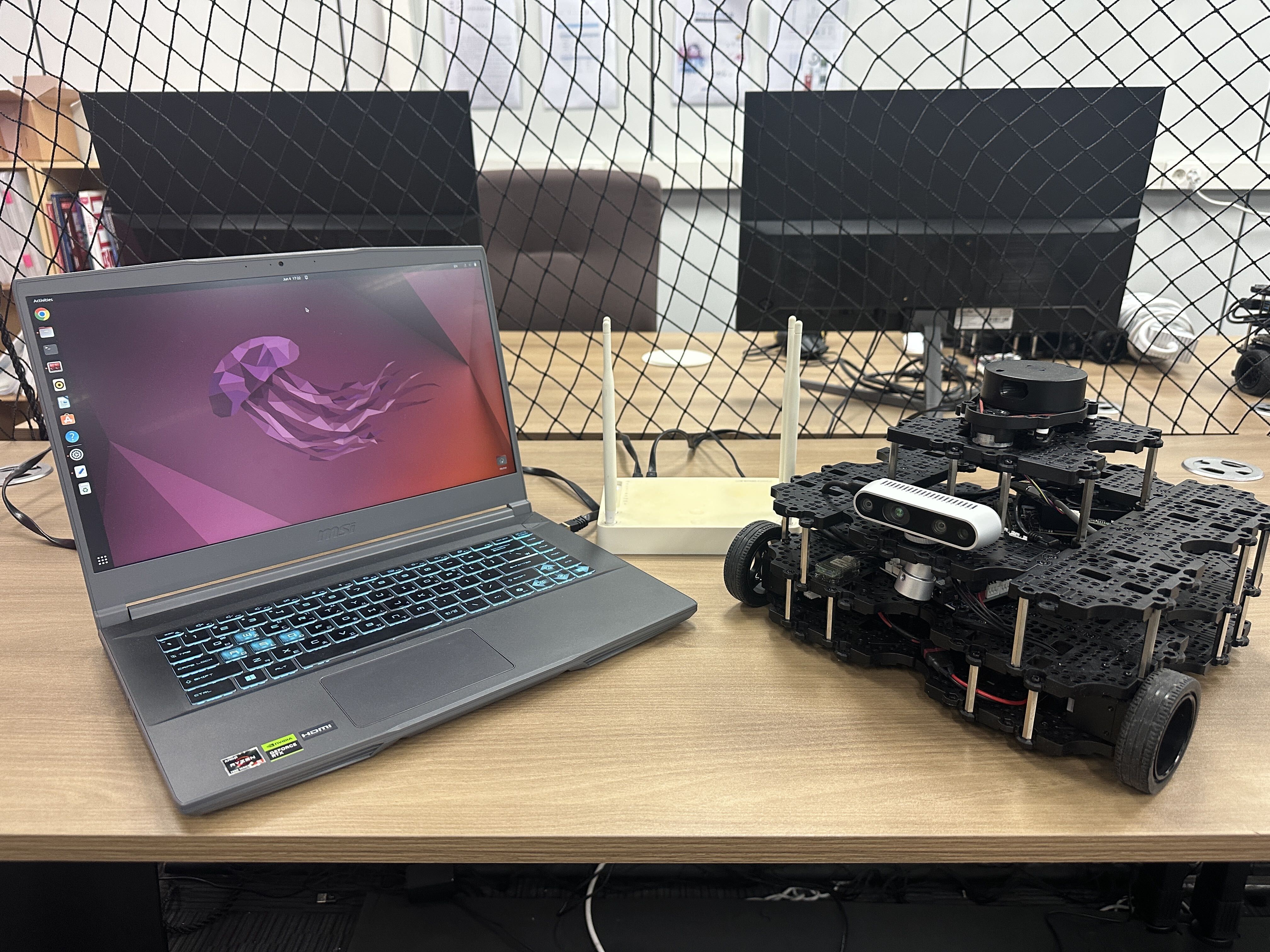}
    \caption{Laptop and Jetson AMR}
    \label{fig:amr-jetson}
  \end{subfigure}
  \hfill
  \begin{subfigure}{0.49\linewidth}
    \includegraphics[width=\linewidth]{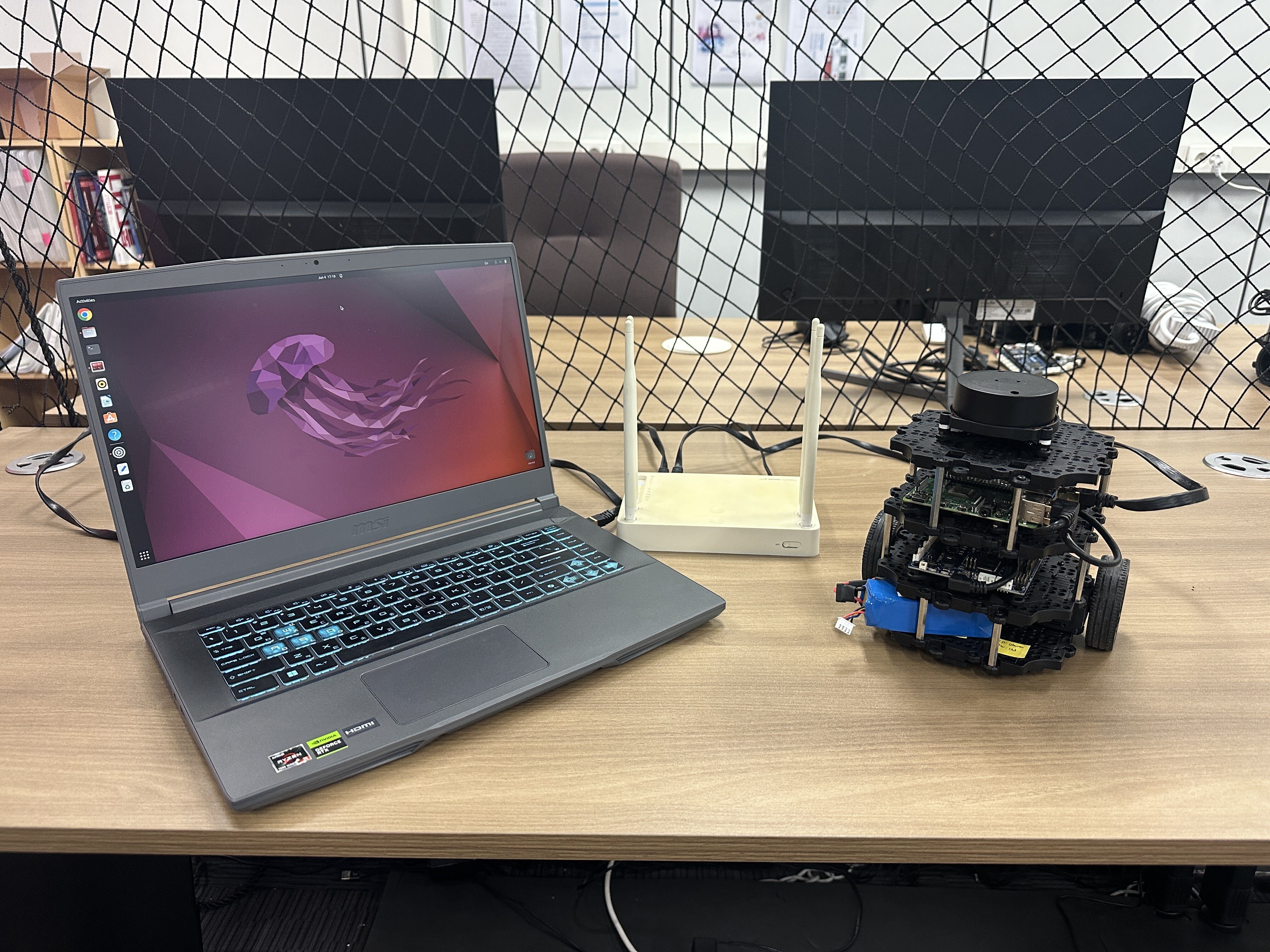}
    \caption{Laptop and Raspberry Pi AMR}
    \label{fig:amr-rpi}
  \end{subfigure}
  \caption{Embedded receivers in Experiment~2 (resource overhead): (a) an NVIDIA Jetson and (b) a Raspberry Pi~4B, each mounted on an AMR.}
  \label{fig:platforms-photo}
  \Description{Photographs of the embedded receivers, an NVIDIA Jetson and a Raspberry Pi 4B, each mounted on an autonomous mobile robot.}
\vspace{-1em}
\end{figure}

\looseness=-1 A DataReader-based tool receives the same stream as the original subscriber once more and processes it through the ROS~2 subscription path with deserialization and QoS handling, so the observer process's CPU and memory cost grows as the message rate increases.
ros2probe, in contrast, is a passive capture that creates no DDS subscriber, and should therefore show lower resource overhead under the same workload.
As Figure~\ref{fig:disc-topo}(a) shows, we verify this in a one-way setup where a fixed publisher laptop transmits over GbE to each of three receiver platforms: a laptop, a Raspberry Pi 4B, and an NVIDIA Jetson. Figure~\ref{fig:platforms-photo} shows the latter two mounted on autonomous mobile robots~(AMR).
The workloads are those in Table~\ref{tab:stress}.
The payload is fixed and only the publication rate is increased, so the comparison targets the per-message processing cost rather than network saturation.
All three receiver platforms fix the CPU governor to performance, and the Jetson is set to its maximum-performance mode.
For rate measurement we compare baseline, rp hz, and ros2 hz, and for recording rp bag and rosbag2, measuring the observer process's CPU utilization and PSS memory (\texttt{/proc/<pid>/smaps_rollup}) under each condition.
The original subscriber's drop rate is used only as a validity check that the workload stayed healthy during the measurement.
Rate conditions are repeated 10 times per platform $\times$ scenario $\times$ condition, and recording conditions are likewise repeated 10 times, except that the Raspberry Pi could not run the ST1000 recording because its SD-card write throughput could not sustain that publication rate.

\begin{figure*}[tbp]
  \centering
  \includegraphics[width=0.85\textwidth]{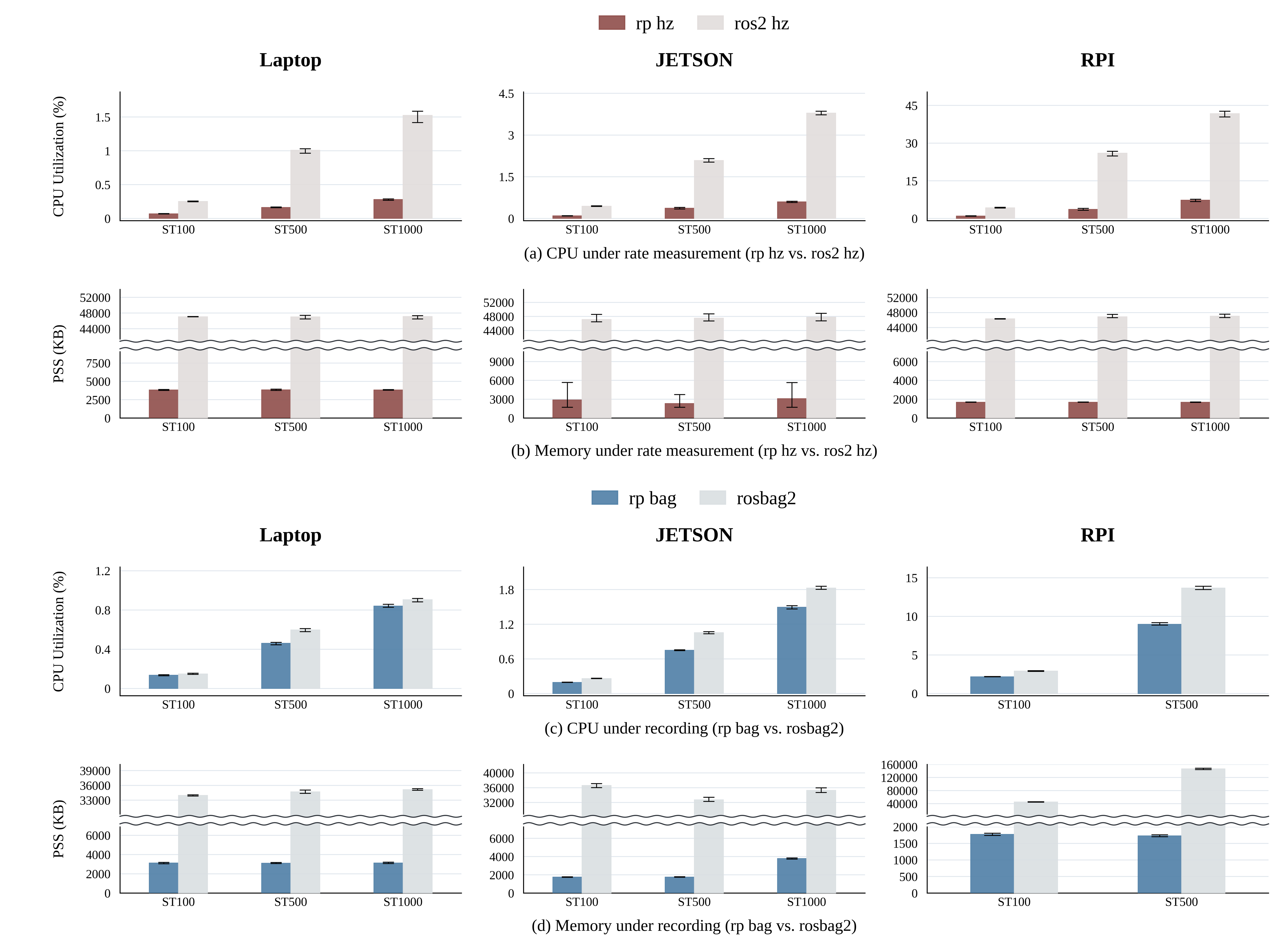}
  \caption{Observer-process resource overhead. (a) CPU and (b) memory under rate measurement (rp hz vs.\ ros2 hz), and (c) CPU and (d) memory under recording (rp bag vs.\ rosbag2). CPU is normalized to total cores, and the Raspberry Pi ST1000 recording is omitted (SD-card write limit).}
  \label{fig:exp3}
  \Description{Bar charts of observer-process CPU and memory under rate measurement and under recording, across laptop, Jetson, and Raspberry Pi platforms, showing ros2probe consuming far fewer resources than the ros2 tools.}
\vspace{-1em}
\end{figure*}

\looseness=-1 As Figure~\ref{fig:exp3} shows, in the rate measurement rp hz used far less observer-process CPU and memory than ros2 hz across all three platforms and all three rates.
The memory gap is especially large and platform-independent. ros2 hz deserializes the messages received via its DataReader through the DDS stack and maintains history buffers and QoS state, occupying about 46 MB on every platform, whereas rp hz only parses RTPS headers and does not deserialize the payload, staying at the capture-buffer-and-GID-map level of 1.7--3.9 MB (about 28$\times$ on the Raspberry Pi 4B).

\looseness=-1 CPU was also about 5--7$\times$ lower for rp hz.
ros2 hz incurs per-message deserialization and callback cost, so its CPU rises in proportion to the rate, whereas rp hz incurs only per-packet header-parsing cost and rises gently as the rate grows.
This gap widens from mere efficiency to feasibility as the rate increases.
On the most constrained Raspberry Pi 4B, ros2 hz's CPU reached 26\% of the four-core capacity at ST500 (one fully occupied core) and 42\% at ST1000 (1.68 cores), whereas rp hz stayed at 3.7\% and 7.3\%.
Meanwhile the original subscriber's drop rate was 0\% on every platform at ST100 and ST500, confirming that the resource figures above were measured while the workload was operating normally.
At ST1000, by contrast, the ros2 tools dropped about 8.6\% of messages on the laptop and Jetson and 16.9\% on the Raspberry Pi, as the observer's extra copy exceeded the link capacity.
Because they then process fewer packets than the offered rate, their CPU at ST1000 is somewhat suppressed, so the true gap to rp hz is even wider than the measured ratio.

\looseness=-1 rp is not free either. Even at this extreme its CPU and memory, far below the ros2 tools, are not zero.
This confirms the resource manifestation of the probe effect from Section~\ref{section1}, where a DataReader-based tool overloads resource-constrained embedded platforms, while ros2probe leaves enough headroom to run practically here.

\looseness=-1 The same trend holds for recording. rp bag's recorder is lighter than rosbag2's.
On the laptop, whose storage is fast enough, the rosbag2 recorder occupied about 35 MB while rp bag used only about 3 MB.
However, on the Raspberry Pi 4B the SD-card write throughput could not keep up with the publication rate, so at ST500 the messages waiting to be written piled up in memory and inflated the recorder memory abnormally, and the ST1000 recording could not be run because of this limit.
On embedded platforms the recording bottleneck is the storage write speed, a constraint separate from the probe effect.

\subsection{Observation Fidelity}\label{subsec:eval-fidelity}
A probe-effect-free observer is useful only if what it records matches what the subscriber actually received.
A DDS-domain observer is served by a separate unicast stream, so under network loss it loses a different set of messages than the subscriber, whereas ros2probe reads the same wire frames delivered to the subscriber and therefore sees the same losses.
As Figure~\ref{fig:disc-topo}(a) shows, we test this on a one-way publisher-to-subscriber link, injecting 0\%, 10\%, and 20\% packet loss on the path with \texttt{tc netem} under BEST_EFFORT QoS.
At each loss level we record the topic with rosbag2 and with rp bag, ten runs each, and compare each recorded set against the messages the subscriber actually received, matched by sequence number.
Table~\ref{tab:fidelity} reports each tool's drop rate and the recall of its reported losses against the subscriber's actual losses, i.e., the fraction of the subscriber's losses the tool also reported.

\begin{table}[tbp]
\centering
\caption{Observation fidelity under injected loss: recall of each tool's reported losses against the subscriber's actual losses (the fraction of the subscriber's losses the tool also reports).}
\label{tab:fidelity}
\footnotesize
\begin{tabular*}{\columnwidth}{@{\extracolsep{\fill}}llrrr}
\toprule
Loss & Tool & Sub.\ drop & Obs.\ drop & Recall \\
\midrule
10\% & rosbag2 & 9.2\% & 11.1\% & 0.09 \\
     & rp bag  & 9.4\% & 9.4\%  & 1.00 \\
20\% & rosbag2 & 20.9\% & 21.0\% & 0.22 \\
     & rp bag  & 19.9\% & 19.9\% & 1.00 \\
\bottomrule
\end{tabular*}
\vspace{-1em}
\end{table}

\looseness=-1 Both tools report nearly the same aggregate drop rate as the subscriber (at 10\% loss, 9.2\% for the subscriber versus 11.1\% for rosbag2 and 9.4\% for rp bag), so the drop rate alone hides the problem.
Which messages were lost, however, diverges sharply.
rp bag reports exactly the subscriber's losses. Its recall is 1.0 at every loss level, so the loss it reports is precisely the subscriber's loss.
rosbag2, in contrast, catches almost none of them. Its recall is only 0.09 (0.22) at 10\% (20\%) loss, because its separate unicast stream drops independently of the subscriber's.
This recall merely tracks the loss rate. For losses independent of the subscriber's, the expected recall equals the observer's own drop rate, so rosbag2's agreement is only coincidental and grows with the loss rate, whereas rp bag reads the same wire and is exact at any loss rate.
This confirms the fidelity manifestation of the probe effect from Section~\ref{section1}. A DDS-domain observer's path is statistically independent of the subscriber's and cannot report which messages the subscriber actually missed, whereas ros2probe, reading the subscriber's own wire traffic, reports loss with full fidelity.

\subsection{Probe Effect}\label{subsec:eval-probe}

\begin{table}[t]
\centering
\caption{Probe-effect evaluation workloads.}
\label{tab:workloads}
\footnotesize
\begin{tabular}{llrrr}
\toprule
Scenario & Topic & Payload & Rate & Est.\ bandwidth \\
\midrule
S1 & \texttt{/imu} & 320 B & 200 Hz & $\sim$0.72 Mbps \\
S2 & \texttt{/scan} & 4.3 KB & 40 Hz & $\sim$1.4 Mbps \\
S3 & \texttt{/points} & 644 KB & 20 Hz & $\sim$103 Mbps \\
S4 & \texttt{/points} & 2.72 MB & 20 Hz & $\sim$435 Mbps \\
S5 & \texttt{/image_raw/compressed} & 150 KB & 30 Hz & $\sim$36 Mbps \\
S6 & \texttt{/depth/image_raw} & 600 KB & 30 Hz & $\sim$147 Mbps \\
S7 & composite (Table~\ref{tab:s7}) & - & - & $\sim$830 Mbps \\
\bottomrule
\end{tabular}
\end{table}

\begin{table}[t]
\centering
\caption{Composition of the S7 composite workload.}
\label{tab:s7}
\footnotesize
\begin{tabular*}{\columnwidth}{@{\extracolsep{\fill}}lrrr@{}}
\toprule
Component topic & Payload & Rate & Est.\ bandwidth \\
\midrule
\texttt{/cmd_vel} & 72 B & 20 Hz & $\sim$0.03 Mbps \\
\texttt{/imu} & 320 B & 200 Hz & $\sim$0.72 Mbps \\
\texttt{/points/front} & 2.72 MB & 20 Hz & $\sim$435 Mbps \\
\texttt{/points/rear} & 644 KB & 20 Hz & $\sim$103 Mbps \\
\texttt{/camera/front/compressed} & 150 KB & 30 Hz & $\sim$36 Mbps \\
\texttt{/camera/left/compressed} & 150 KB & 30 Hz & $\sim$36 Mbps \\
\texttt{/camera/right/compressed} & 150 KB & 30 Hz & $\sim$36 Mbps \\
\texttt{/camera/rear/compressed} & 150 KB & 30 Hz & $\sim$36 Mbps \\
\texttt{/depth/image_raw} & 600 KB & 30 Hz & $\sim$147 Mbps \\
\midrule
\textbf{Total} & & & \textbf{$\sim$830 Mbps} \\
\bottomrule
\end{tabular*}
\end{table}

ros2probe creates no DDS DataReader, so the publisher sends nothing extra to the observer. Enabling the observer should therefore not increase the traffic received by the original subscriber.
As Figure~\ref{fig:disc-topo}(a) shows, we verify this in a one-way setup between two identical laptops, listed in Table~\ref{tab:platforms}, one as the publisher and the other as the subscriber-and-observer host.
The workloads are the seven scenarios of Table~\ref{tab:workloads}, sweeping the bandwidth from a small sensor message (S1) to a composite autonomous-driving workload near GbE saturation (S7).
The severity of the probe effect, however, depends on more than the workload. The QoS reliability setting and the link medium both shape it.
Under RELIABLE QoS the publisher adds Heartbeat/AckNack control traffic and retransmissions for each matched reader on top of the duplicated data stream, and a wireless link offers lower, more variable capacity, so saturation arrives at a lower offered load~\cite{lee2025optimizing}.
We therefore sweep the full $2\times2$ matrix of \{BEST_EFFORT, RELIABLE\} QoS and \{wired, wireless\} link, reporting all four cells below.
The baseline is the condition with no observer.
We compare two functions: rate measurement, which reports a topic's publication rate, and recording, which saves its messages to an MCAP file.
For rate measurement we pair the ROS~2 standard tool \texttt{ros2 topic hz} (ros2 hz) with ros2probe's \texttt{rp topic hz} (rp hz), and for recording the standard \texttt{ros2 bag record} (rosbag2) with \texttt{rp bag record} (rp bag).
For rate measurement the observer watches a single target topic, \texttt{/points/front} in the composite scenario S7, whereas for recording the observer always records every topic in the workload via the recorders' \texttt{-{}-all} option. In the single-topic scenarios S1--S6 the two coincide, since only S7 carries multiple topics. Each condition is compared against the baseline.
Each run launches the observer first so that it attaches before the publisher, then the subscriber and the NIC/CPU/memory samplers, and finally the publisher.
The 60-second window starts at the subscriber's first received message, excluding the discovery and warm-up phase. After each run the processes are cleaned up and the next run starts 10 seconds later.
Every condition is repeated 10 times.
The received bandwidth (BW) is obtained by reading the observation host NIC's \texttt{/proc/net/dev} received-byte counter every second to compute the per-second rate, then averaging these per-second values over the 60-second window.
The original subscriber's drop rate is computed by comparing the messages it received in the same window against the expected count (rate $\times$ 60 s).

\begin{table*}[t!]
\centering
\caption{Received bandwidth (BW, Mbps) and original-subscriber drop rate (Loss, \%) per scenario.}
\label{tab:rxdrop}
\footnotesize
\setlength{\tabcolsep}{3pt}
\begin{tabular*}{\textwidth}{@{\extracolsep{\fill}}ll*{5}{rr}@{}}
\toprule
& & \multicolumn{2}{c}{Baseline} & \multicolumn{2}{c}{rp hz} & \multicolumn{2}{c}{ros2 hz} & \multicolumn{2}{c}{rp bag} & \multicolumn{2}{c}{rosbag2} \\
\cmidrule(lr){3-4}\cmidrule(lr){5-6}\cmidrule(lr){7-8}\cmidrule(lr){9-10}\cmidrule(lr){11-12}
Link / QoS & Sc. & BW & Loss & BW & Loss & BW & Loss & BW & Loss & BW & Loss \\
\midrule
\multirow{7}{*}{\shortstack[l]{wired\\BEST_EFFORT}}
 & S1 & 0.70 & 0 & 0.70 & 0 & 1.35 & 0 & 0.70 & 0 & 1.35 & 0 \\
 & S2 & 1.46 & 0 & 1.46 & 0 & 2.94 & 0 & 1.46 & 0 & 2.94 & 0 \\
 & S3 & 107  & 0 & 107  & 0 & 216  & 0 & 107  & 0 & 216  & 0 \\
 & S4 & 466  & 0 & 466  & 0 & 934  & 0 & 466  & 0 & 934  & 0 \\
 & S5 & 37.5 & 0 & 37.5 & 0 & 75.3 & 0 & 37.5 & 0 & 75.3 & 0 \\
 & S6 & 150  & 0 & 150  & 0 & 301  & 0 & 150  & 0 & 301  & 0 \\
 & S7 & 874  & 0 & 874  & 0 & 981  & 38.5 & 874  & 0 & 984  & 75.5 \\
\midrule
\multirow{7}{*}{\shortstack[l]{wired\\RELIABLE}}
 & S1 & 0.75 & 0 & 0.75 & 0 & 1.45 & 0 & 0.75 & 0 & 1.45 & 0 \\
 & S2 & 1.47 & 0 & 1.47 & 0 & 2.94 & 0 & 1.47 & 0 & 2.94 & 0 \\
 & S3 & 108  & 0 & 108  & 0 & 216  & 0 & 108  & 0 & 216  & 0 \\
 & S4 & 490  & 0 & 490  & 0 & 978  & 12.3 & 490  & 0 & 979  & 19.3 \\
 & S5 & 37.6 & 0 & 37.6 & 0 & 75.4 & 0 & 37.6 & 0 & 75.4 & 0 \\
 & S6 & 150  & 0 & 150  & 0 & 301  & 0 & 150  & 0 & 301  & 0 \\
 & S7 & 976  & 24.0 & 976  & 25.5 & 984  & 42.5 & 976  & 19.5 & 984  & 79.3 \\
\midrule
\multirow{7}{*}{\shortstack[l]{wireless\\BEST_EFFORT}}
 & S1 & 0.70 & 0 & 0.70 & 0 & 1.35 & 0 & 0.70 & 0 & 1.35 & 0 \\
 & S2 & 1.46 & 0 & 1.46 & 0 & 2.94 & 0 & 1.46 & 0 & 2.94 & 0 \\
 & S3 & 107  & 0 & 107  & 0 & 216  & 0 & 107  & 0 & 216  & 0 \\
 & S4 & 336  & 27.8 & 336  & 27.8 & 336  & 64.3 & 336  & 28.8 & 326  & 63.8 \\
 & S5 & 37.5 & 0 & 37.5 & 0 & 75.4 & 0 & 37.5 & 0 & 75.4 & 0 \\
 & S6 & 150  & 0 & 150  & 0 & 301  & 0 & 150  & 0 & 300  & 0 \\
 & S7 & 427  & 98.5 & 435  & 97.7 & 432  & 99.7 & 435  & 98.5 & 438  & 98.7 \\
\midrule
\multirow{7}{*}{\shortstack[l]{wireless\\RELIABLE}}
 & S1 & 0.75 & 0 & 0.75 & 0 & 1.45 & 2.9 & 0.75 & 0 & 1.45 & 2.9 \\
 & S2 & 1.47 & 0 & 1.47 & 0 & 2.95 & 0 & 1.47 & 0 & 2.95 & 0 \\
 & S3 & 148  & 0 & 148  & 0 & 180  & 18.5 & 148  & 0 & 180  & 19.2 \\
 & S4 & 163  & 67 & 163  & 69.5 & 165  & 82.5 & 163  & 67 & 170  & 83.4 \\
 & S5 & 37.6 & 0 & 37.6 & 0 & 75.5 & 0 & 37.6 & 0 & 75.5 & 0 \\
 & S6 & 161  & 1.8 & 161  & 1.3 & 178  & 41.8 & 161  & 3.1 & 178  & 41.8 \\
 & S7 & 154  & 95.2 & 154  & 95 & 156  & 95 & 154  & 95.3 & 155  & 95.8 \\
\bottomrule
\end{tabular*}
\end{table*}

As Table~\ref{tab:rxdrop} shows, across every cell of the matrix rp hz and rp bag receive essentially the baseline bandwidth, because they only read copies already on the wire and add no transmission of their own.
The DDS tools instead pull a second unicast stream from the publisher, roughly doubling the received bandwidth wherever the link still has headroom, for example on wired S1--S6 and on the lower-bandwidth wireless scenarios.
Once the offered load plus this extra copy exceeds the link capacity, the received bandwidth saturates at the link ceiling and the surplus demand turns into loss, which the drop analysis below examines.

On the wired GbE link under BEST_EFFORT QoS, the duplicated stream fits below the link until the composite scenario S7.
There ros2 hz observes only \texttt{/points/front}, so the publisher sends one extra copy of it ($\sim$435 Mbps); the link must then carry about 1.27 Gbps ($\approx 830+435$) against the $\sim$0.94 Gbps usable on 1 GbE, and the subscriber loses 38.5\% of its messages. rosbag2 instead records every topic, so the publisher duplicates the entire $\sim$830 Mbps workload and the loss climbs to 75.5\%.
rp, creating no copy, stays at 0\%, as Table~\ref{tab:rxdrop} reports.
RELIABLE QoS makes the effect appear earlier and grow stronger, because Heartbeat/AckNack control traffic and retransmissions add to the offered load.
It already shows at S4 (loss-free for every condition under BEST_EFFORT), where the DDS tools drop 12.3\% and 19.3\%, and at the saturating S7 retransmissions eat into delivery even without an observer (baseline 24.0\%) while ros2 hz pushes the loss to 42.5\% and rosbag2, recording every topic, to 79.3\%.
rp tracks the baseline throughout.

According to Table~\ref{tab:rxdrop}, the wireless link offers far less, and more variable, capacity than wired GbE, so saturation, and with it the probe effect, arrives at a much lower offered load.
It also behaves differently under overload. Unlike the wired link, whose received bandwidth saturates at the link ceiling, the contended and lossy wireless medium cannot sustain its peak, so at saturation the received bandwidth collapses far below the offered load, to only $\sim$150--165 Mbps under RELIABLE at S4, S6, and S7, instead of holding at a ceiling.
Already at S4 the workload alone fills the link, so the baseline itself loses 27.8\% (BEST_EFFORT) and 67\% (RELIABLE), and the DDS tools' extra copy worsens this to 64.3\%/63.8\% and 82.5\%/83.4\%.
By S7 the link is fully saturated for every condition ($\sim$98\% under BEST_EFFORT, $\sim$95\% under RELIABLE).
Where headroom remains, the DDS observer still induces loss that the baseline does not. Under RELIABLE, S3 is loss-free and S6 nearly so (1.8\%) at baseline, yet they drop 18.5--19.2\% and 41.8\% once a DDS tool joins, while rp again matches the baseline in every wireless cell.
The probe effect, confined to near-saturation on wired GbE, thus spreads across many more scenarios on the thinner wireless link, whereas rp stays transparent on both.

Across the whole $2\times2$ matrix and the full workload sweep, rp is the only tool whose received bandwidth and drop rate stay within measurement error of the baseline. It neither adds traffic nor perturbs delivery, and the small residual differences from the baseline reflect run-to-run variance rather than any effect of observation.
The DDS tools, by contrast, distort the system exactly when observation matters most (under high load, reliable delivery, or a constrained link), and the loss they cause falls on the original subscriber's path, absent from the tool's own output, so the danger stays hidden until it is already severe.
This empirically confirms the structural limitation of Section~\ref{subsec:probe-effect}. Observation from inside the DDS domain can never be read-only on the wire.

\section{Conclusion}\label{section6}
This paper formulates the probe effect in ROS~2 observation not as a flaw of individual tools but as a structural consequence of the DDS pub/sub design, in which only a DataReader matched through discovery can receive data, and proposes ros2probe, an observation system that circumvents this limitation.
ros2probe reconstructs the ROS~2 topic graph and per-topic metrics by passively capturing and interpreting RTPS wire traffic outside the DDS domain.
By selecting only the topics of interest in the kernel and not deserializing payloads, it provides real-time observation at low overhead without participating in the domain.

\looseness=-1 In the evaluation, ros2probe did not affect the incoming traffic or the original subscriber's drop up to workloads near GbE saturation, left no trace in the discovery graph, and reduced the observer process's CPU by about 7$\times$ and memory by about 28$\times$ relative to existing tools, operating even on embedded platforms where existing tools are overloaded.
ros2probe thus simultaneously provides the four properties that no existing tool offers together: no probe effect, DDS-implementation independence, real-time semantic information, and observation cost that scales with the observed topics rather than the total traffic.
It can be dropped in immediately with output and recording formats compatible with existing tools.
Since ros2probe currently targets RTPS traffic that appears on the wire, we leave support for non-DDS transports such as Zenoh to future work.

\bibliographystyle{ACM-Reference-Format}
\bibliography{references}


\end{document}